\documentclass{article}
\usepackage{ijcai25}

\usepackage{times}
\usepackage{soul}
\usepackage{url}
\usepackage[hidelinks]{hyperref}
\usepackage[utf8]{inputenc}
\usepackage[small]{caption}
\usepackage{graphicx}
\usepackage{amsmath}
\usepackage{amsthm}
\usepackage{booktabs}
\usepackage{algorithm}
\usepackage{algorithmic}
\usepackage[switch]{lineno}

\usepackage{xcolor}
\usepackage{amsmath,amssymb,amsfonts}
\usepackage{algorithmic}
\usepackage{algorithm}
\usepackage{algorithmic}
\usepackage{graphicx}
\usepackage{subfigure}
\usepackage{tabularx}
\usepackage{booktabs}
\usepackage{multirow}
\usepackage[title]{appendix} 

\newcommand{\kevin}[1]{{\color{violet} (Kevin: #1)}}

\newcommand{\guowen}[1]{{\color{orange} (Guowen: #1)}}

\newcommand{\method}{\textsc{Adell}}
\newcommand{\Mlang}{\mathcal{M}_{l}}
\newcommand{\Mclass}{\mathcal{M}_{c}}
\newcommand{\sfe}{{S_\mathit{fe}}}
\newcommand{\sse}{{S_\mathit{se}}}

\usepackage{comment}

\title{Adversarial Demonstration Learning for Low-resource NER \\ Using Dual Similarity}

\author{
Guowen Yuan$^1$
\and
Tien-Hsuan Wu$^1$\and
Lianghao Xia$^{1}$\And
Ben Kao$^1$\\
\affiliations
$^1$The University of Hong Kong\\
\emails
csgwy@connect.hku.hk,
thwu@cs.hku.hk,
aka\_xia@foxmail.com,
kao@cs.hku.hk
}

\begin{document}

\maketitle

\begin{abstract}

We study the problem of named entity recognition (NER) based on demonstration learning in low-resource scenarios.
We identify two issues in {\it demonstration construction} and {\it model training}.
Firstly, existing methods for selecting demonstration examples primarily rely on semantic similarity; We show that
feature similarity can provide significant performance improvement. 
Secondly, we show that the NER tagger's ability to reference demonstration examples is generally inadequate.
We propose a demonstration and training approach that effectively addresses these issues.
For the first issue, we propose to select examples by dual similarity, which comprises both semantic similarity and feature similarity.
For the second issue, we propose to train an NER model with adversarial demonstration such that the model is forced to refer to the demonstrations when performing the tagging task.
We conduct comprehensive experiments in low-resource NER tasks, and the results demonstrate that our method outperforms a range of methods.

\end{abstract}

\section{Introduction}

Named Entity Recognition (NER) is a crucial technique in the field of natural language processing (NLP) that involves identifying and categorizing named entities within unstructured text. The extracted information serves various applications such as the construction of knowledge graphs, information retrieval, and question answering. Developing an accurate NER model requires a substantial amount of labeled data. However, the process of acquiring extensive annotations is both time-consuming and resource-intensive. For instance, in \cite{wu2020JURIX}, it is reported that a task of annotating  4,000 court judgments took 11 workers with legal training six months to complete.
In such cases, NER model training is limited by relatively small amounts of labeled data, for which we describe the situation as ``low resource''.
In low-resource scenarios, the significance of effective NER techniques is amplified, as it directly influences the feasibility of applying NER in real-world tasks where data is scarce.

\begin{figure}[!]
\centering
\includegraphics[width=\columnwidth]{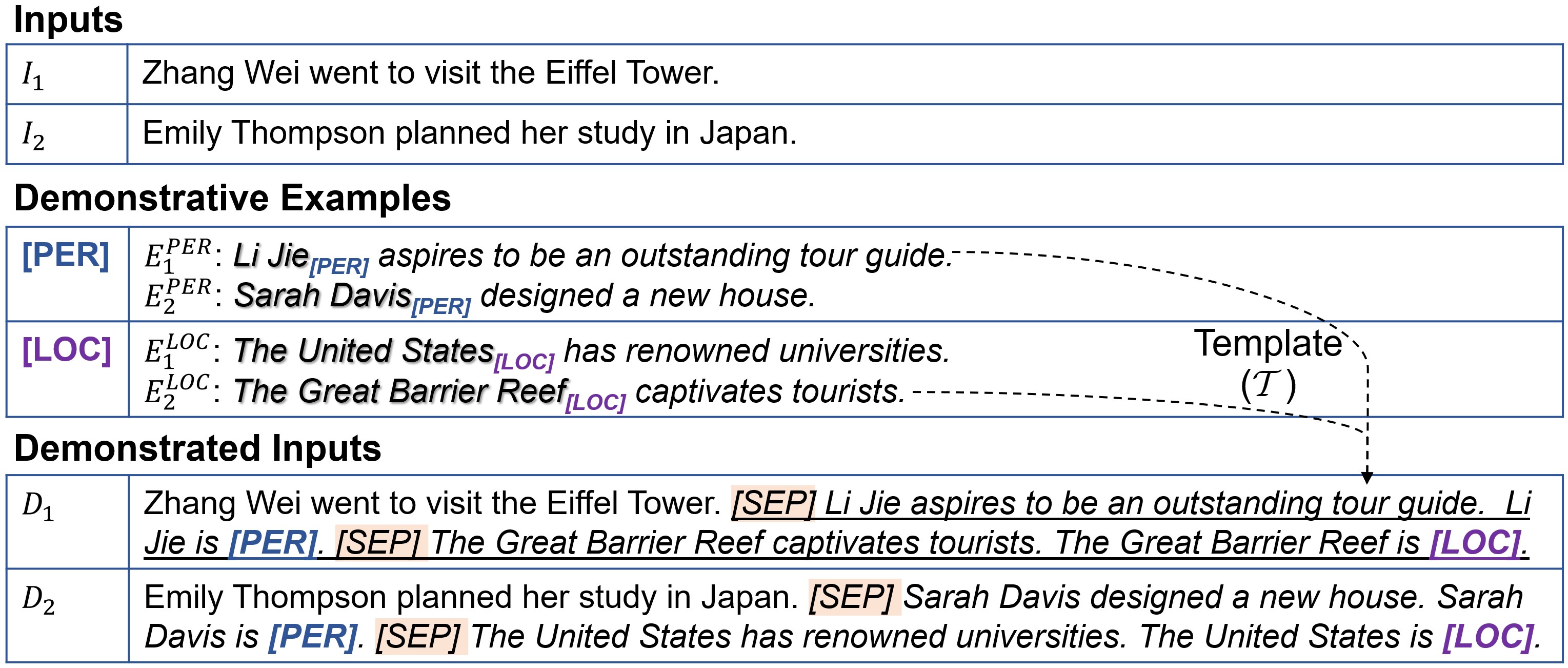}
\caption{An illustrative example of demonstration}
\label{fig:ExampleDynamicDemanstration}
\end{figure}

Previous works~\cite{gao2021making,liu2023pre} have shown that demonstration learning improves the performance of pre-trained language models, and~\cite{lee2022good} specifically demonstrated its effectiveness in low-resource NER settings.
The idea of demonstration learning is to 
pair
an input data instance (text to be NER-tagged) with a {\it demonstration}, which is a small
number of labeled examples. Together, they form a ``data with demonstration'' (or ``demonstrated'') input. 
Figure~\ref{fig:ExampleDynamicDemanstration} illustrates such input created by a {\it demonstration incorporator}.
Specifically, for each input instance, its demonstration includes a labeled example for each label category.
In Figure~\ref{fig:ExampleDynamicDemanstration}, the input $I_1$ is demonstrated by similar examples, such as $E_1^\mathit{PER}$ for the [PER] category (both mention English transliteration of Chinese names) and  $E_2^\mathit{LOC}$ for the [LOC] category (both mention tourist spots).
The demonstrated input $D_1$ is constructed by concatenating the input $I_1$ together with the examples $E_1^\mathit{PER}$ and $E_2^\mathit{LOC}$. 
(A special token [SEP] is inserted in between the concatenating pieces.)
Demonstrated inputs are used in both model training and application. 
In training, input instances are labeled; In application, the input instance is previously unseen data, whose labels are to be determined by the trained NER model.

Our objective is to study how demonstration learning can be most effectively applied in a low-resource situation.
In particular, we put forward two principles in demonstration learning and study how those could be achieved:
(Attention) Since demonstration examples help guide the NER labeling, the NER model should pay enough attention to the 
given examples. 
(Similarity) The selected demonstration examples should be most similar to the input instance so that they provide the best guidance. 
By studying existing demonstration learning approaches, we observe that existing methods could be improved in two important aspects:

\noindent\textbf{[Feature Similarity]} Existing methods select demonstration examples based mainly on semantic similarity
(e.g., using text embeddings).
However, {\it feature similarity} should also be considered for composing the best demonstration.
Feature similarity refers to the similarity between the feature labels (such as entity types in NER tasks)
of the input instances and those of candidate demonstration examples. 
Our investigation shows that the two kinds of similarity are complementary and thus should be integrated. 
To utilize feature similarity, 
we implement a {\it feature similarity predictor} (Section~\ref{sec:dual_similarity}) in the demonstration incorporator to help select the best demonstration examples.

\noindent\textbf{[Demo Attentiveness]} In a low-resource scenario, existing demonstration learning models tend to rely 
more on the training set examples and disregard the demonstration examples. We will show that helping the model recognize the importance of demonstrations during training can lead to a better performance.
Specifically, we propose {\it Adversarial Demonstration Training (ADT)} (Section~\ref{sec:ner_training_with_game}). 
Our technique is to perturb the annotation rules during training such that the NER model must refer to the rules embedded in the demonstration examples and mimics these rules to annotate the input text, while providing correct rules during application to ensure accurate annotations. 
This helps train the model to pay attention to demonstrations. 
Our experiment results show that ADT can substantially improve NER model performance.

\section{Related Work}
\label{sec:related}









Low-resource NER is a challenging task that arises when there is a scarcity of labeled data, often encountered in the initial stage of an NER task. 
Numerous methods have been proposed to alleviate the problem of low-resource NER. 
Data augmentation~\cite{dai2020analysis,yaseen2021data} includes techniques such as label-wise token replacement, synonym replacement, mention replacement, shuffle within segments, and translation.
By modifying the data collected in the training set, a larger, augmented dataset is created for better NER model training.

In addition to data augmentation, some methods utilize prior knowledge, such as metric-based learning, implicit prompt tuning, and explicit prompt learning. 
Metric-based learning \cite{snell2017prototypical,fritzler2019few,yang2020simple} aims to learn a metric to accurately quantify the distance or similarity between all pairs of examples. For example, \cite{yang2020simple} assigns to a given unlabeled token, the tag of the token's nearest neighbor token (NNShot). Another method (StructShot) uses the Viterbi decoder to infer the label. In this method, a matrix is used to capture the transition probabilities of the labels of neighboring tokens in the training set.
Implicit prompts \cite{liu2021gpt,liu2021p} are designed to distill complex knowledge that cannot be easily expressed in a few phrases, allowing targeted fine-tuning to be done without modifying all parameters of the large language model. In P-tuning~\cite{liu2021gpt}, the tuning is done on the input layer of the prompt, whereas P-tuning v2~\cite{liu2021p} tunes prompts in the input and all hidden layers.
Numerous methods have been proposed for utilizing explicit prior knowledge, such as template-based NER \cite{cui2021template}, template-free NER (EntLM \cite{ma_template_free_2022}, DualBERT \cite{ma2022label}) and demonstration learning \cite{wang_training_2022,gao2021making,lee2022good}. 

Demonstration learning has been shown to provide good performance with task-specific knowledge. There are studies 
on improving demonstration learning. For example, 
\cite{lee2022good} proposes to select demonstration examples that best improve the labeling w.r.t. a validation set. 
\cite{gao2021making} proposes semantic-based example selection using SBERT~\cite{reimers2019sentence}.
In this paper we further improve demonstration learning with two techniques, namely, feature-based similarity and adversarial demonstration training.

\section{Method}
In this section we formally define the NER problem and describe in detail our method.
\subsection{Definitions}
\label{sec:def}

\textbf{[NER]}
Let $d_i = [w_{i1},...,w_{i|d_i|}]$ be an {\bf input instance} (such as a sentence or a document) formed by a sequence of tokens (words), where $|d_i|$ denotes the number of tokens. 
Let $\mathcal{F}$ = $\{F_1, ..., F_{|\mathcal{F}|}\}$ be a pre-defined set of features. 
Given input $d_i$, the task ({\bf NER-tagging}) is to annotate it by assigning features to certain segments of $d_i$, assuming that each segment is given at most one feature. 
Specifically, NER assigns a feature label $l_{ij} \in \mathcal{F}$ to each token $w_{ij}$ of $d_i$, or the label ``NIL'' if the corresponding token is not part of any feature.
We abstract the annotation made on an input $d_i$ as a sequence of markups $A_i = [m_{i1}, ..., m_{i|A_i|}]$, where $|A_i|$ denotes the number of markups in $A_i$. Each markup $m_{ij}$ has the form $(t_{ij}, f_{ij})$, where $t_{ij}$ is a span of text in $d_i$ and $f_{ij} \in \mathcal{F}$. 
The markup $m_{ij}$ expresses that
``the $j$-th markup in input $d_i$ contains the text $t_{ij}$ and is assigned the feature label $f_{ij}$.''
As an example, in Figure~\ref{fig:ExampleDynamicDemanstration}, the instance
$E_{1}^{\mathit{PER}}$ has one markup, namely, (Li Jie, [PER]).
If the markups of an input $d_i$ are known, we say that $d_i$ is {\it labeled}; otherwise $d_i$ is an {\it unlabeled} input.

\noindent\textbf{[Example and Demonstration]}
We assume a training set $T$ of labeled inputs is available, which are also called {\it labeled examples}. 
A subset $C$ of $T$ is collected as a {\it demo pool} of {\it demonstrative examples}.
Given an input $d_i$, we construct a {\bf demonstration} $X_i$ by selecting some demonstrative examples (from $C$) that are {\it similar} to $d_i$. 
A {\it template function} $\mathcal{T}(X_i)$ converts a demonstration $X_i$ to a string that 
displays the demonstrative examples in $X_i$ and their markups. 
If an input $d_i$ is given a demonstration ($X_i$), then the content, denoted by $[d_i:\mathcal{T}(X_i)]$,
is a {\it demonstrated input}. 
This is illustrated in Figure~\ref{fig:ExampleDynamicDemanstration} where two demonstrative examples,
$E_1^\mathit{PER}$ and $E_2^\mathit{LOC}$, are selected for the input $I_{1}$.
Hence, the demonstration $X_{I_{1}} = [E_1^\mathit{PER}$; $E_2^\mathit{LOC}]$.
The demonstrated input $D_1$ is then obtained by concatenating $I_1$ with the string displayed by
the template function $T(X_{I_{1}})$ as shown in the figure.

\subsection{ADELL}
\label{sec:adell}


\begin{figure*}[!]
    \centering
    \includegraphics[width=0.85\linewidth]{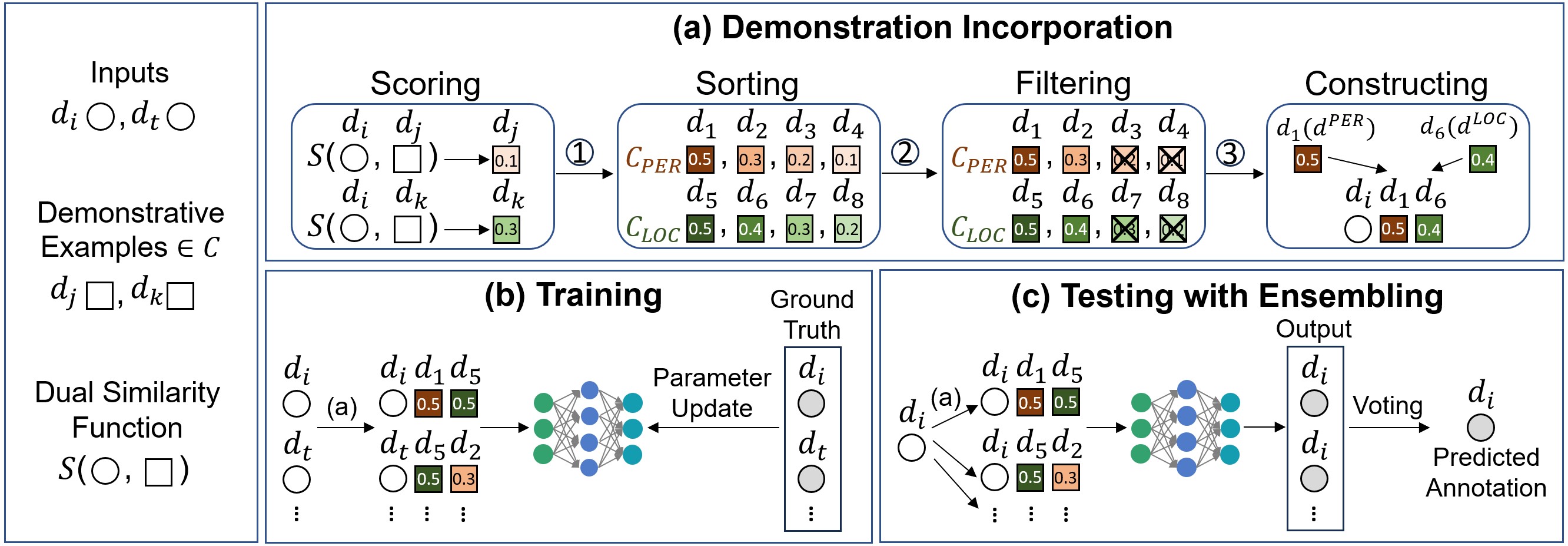}
    \caption{Framework of NER with demonstration.}
    \label{fig:framework_dynamicdemonstration_learning}
\end{figure*}





Figure~\ref{fig:framework_dynamicdemonstration_learning} illustrates the overall framework of our proposed method \method. The framework consists of three modules, namely, {\it Demonstration Incorporator}, {\it Training with Demonstration}, and {\it Application with Demonstration}. In the figure, we use circles ($\bigcirc$) and boxes ($\square$) to represent input and demonstrative examples, respectively. Different colors of the boxes indicate different feature types (e.g., orange for [PER]; green for [LOC]). 
Given an input $d_i$, a box (demonstrative example) with a darker color shade indicates that the example is more 
similar to $d_i$ 
with respect to dual similarity
(as represented by the color). 

\subsubsection{Demonstration Incorporator (DI)}
\label{sec:di}
Given an input instance $d_i$, the purpose of the DI module is to construct a demonstration $X_i$. 
As we previously mentioned, the demonstration $X_i$ consists of a number of (labeled) demonstrative examples 
selected from a demo pool $C$. As shown in~\cite{gao2021making}, it is important that the selected examples are in some way similar to the input $d_i$ in order for demonstration learning to be effective. 
To achieve this goal, \method's DI module carries out a few steps as illustrated in Figure~\ref{fig:framework_dynamicdemonstration_learning}(a).
First, \method\ employs a {\it dual similarity function} $S(d_i, d_j)$ to measure the similarity between an input instance $d_i$ and each demonstrative example $d_j \in C$. 
The dual similarity function considers both semantic similarity and feature similarity. 
(We will further elaborate on the function shortly.)
\method\ first prepares a number of subsets of $C$, one for each feature label.  
Specifically, for each feature $f$ in the feature set $\mathcal{F}$, the subset $C_f \subseteq C$ collects 
all examples $d_j \in C$ that contain feature $f$ in their markups, i.e., 
$C_f = \{d_j \in C | (\cdot, f) \in A_j\}, \forall f \in \mathcal{F}$. (Recall that $A_j$ denotes the annotation of the example $d_j$.) Next, for each $C_f$,  \method\ ranks the examples in $C_f$ in decreasing similarity scores w.r.t. the input instance $d_i$ (i.e., in decreasing $S(d_i, d_j)$, where $d_j \in C_f$). 
Following the idea of {\it example filtering}~\cite{gao2021making}, half of the examples in $C_f$ that have the lowest scores are removed. Then, we randomly select an example $d^f$ from the resulting $C_f$. 
The process is repeated for each $f \in \mathcal{F}$, which results in $|\mathcal{F}|$ demonstrative examples drawn.
These examples are then sorted in decreasing order according to their similarity with the input instance $d_i$ before they
are concatenated to form a demonstration $X_i$.

\subsubsection{Dual Similarity Measure}
\label{sec:dual_similarity}
\begin{figure}[!]
    \centering
    \includegraphics[width=0.47\textwidth]{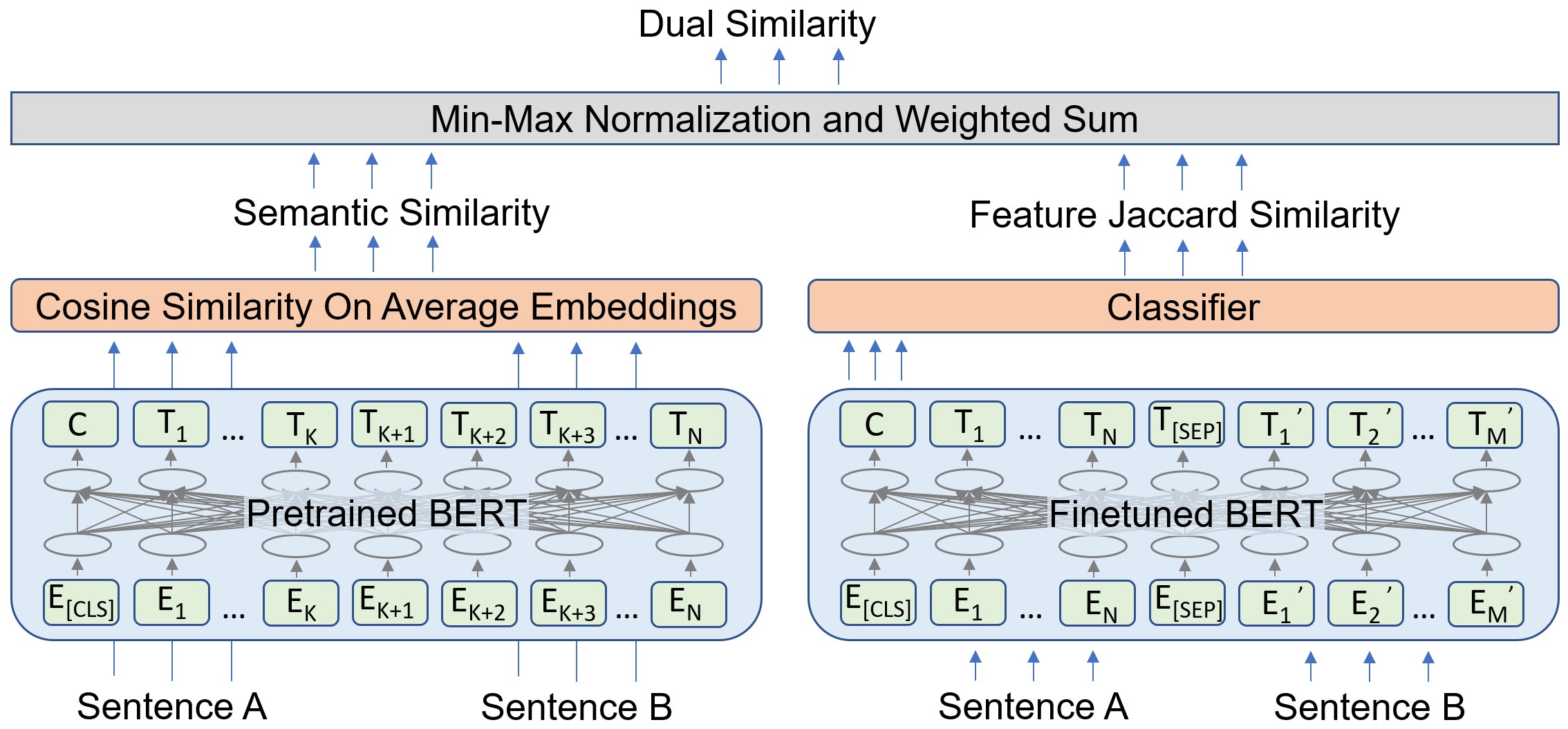}
    \caption{Similarity prediction.}
    \label{fig:SimilarityPredictior}
\end{figure}

There are previous works that discuss how demonstrative examples are selected. However, most of these works consider only semantic similarity.
For example, in~\cite{gao2021making}, similarity between an input instance $d_i$ and a labeled example $d_j$ is 
measured based on the text embeddings (using SBERT~\cite{reimers2019sentence}) of the tokens in $d_i$ and $d_j$. 
We propose to enhance the similarity measure by considering also {\it feature similarity}, i.e., whether $d_i$ and $d_j$
contain similar sets of features. We remark that feature similarity is highly important for demonstration learning because 
the NER model can perform more effectively if it is given demonstrative examples ($d_j$) with similar features as the 
target input instance ($d_i$). 
As we will show with our experiment results (and also demonstrated in~\cite{lee2022good}), there is little correlation between semantic and feature similarities. We therefore propose a dual similarity measure that captures both kinds of similarity. Figure~\ref{fig:SimilarityPredictior} illustrates the idea.

Specifically, given an input $d_i$ and a labeled example $d_j$, the dual similarity is given by
\begin{equation}
\label{eq:dual-sim}
S(d_i, d_j) = \gamma \cdot S_\mathit{fe}(d_i, d_j) + (1-\gamma) \cdot S_\mathit{se}(d_i, d_j),
\end{equation}
where $S_\mathit{fe}()$ and $S_\mathit{se}()$ are the feature similarity and the semantic similarity, respectively, and $\gamma$ is a weighting factor.
The semantic similarity score $S_\mathit{se}(d_i, d_j)$ is given by the cosine similarity of the 
embedding outputs on the token strings of $d_i$ and $d_j$ using the pre-trained BERT sentence encoder
(see left component of Figure~\ref{fig:SimilarityPredictior}).
Since the features contained in $d_i$ can be unknown, 
to measure feature similarity, $S_\mathit{fe}()$, we train a cross-encoder $\mathcal{M}_{c}([d_i;d_j])$ to predict the Jaccard similarity of the features between $d_i$ and $d_j$ using a fine-tuned BERT model
(see right component of Figure~\ref{fig:SimilarityPredictior}).
Given two labeled examples $d_a$ and $d_b$, the {\it feature Jaccard similarity} is calculated by 
\begin{small}
\begin{equation}
\mathit{FJ}(d_a, d_b) = \frac{|\{f | (., f) \in A_a\} \; \cap \; \{f | (., f) \in A_b\}|}{|\{f | (., f) \in A_a\}\;  \cup\;  f | (., f) \in A_b\}|}.
\label{eq:fj}
\end{equation}
\end{small}

\subsubsection{Model Training and Application}
To train an NER model with demonstration learning, we apply the demonstration incorporator DI 
to convert each (labeled) input in the training set into a demonstrated input. 
We then train the NER model using the demonstrated inputs with
StructShot (see Fig.~\ref{fig:framework_dynamicdemonstration_learning}(b)).
In testing/applying the model to an input $d_i$, we apply DI to $d_i$ $k$ times to form $k$ demonstrated inputs. 
The predicted annotation of $d_i$ is obtained via majority voting of the $k$ inputs ensemble (see 
Fig.~\ref{fig:framework_dynamicdemonstration_learning}(c)).

\subsection{Adversarial Demonstration Learning}
\label{sec:ner_training_with_game}
We observe that in traditional demonstration learning, the NER model trained may not pay much attention to the demonstrative examples. 
This significantly weakens the effectiveness of demonstration learning. 
In this section we propose Adversarial Demonstration Learning (ADL) employed by our method \method\ to tackle this issue.

Our NER model is constructed based on StructShot \cite{yang2020simple}, which 
consists of two components: (1) a language model ($\Mlang$) that embeds tokens (of an input string) into their embeddings and (2) a classifier model ($\Mclass$) that considers the token embeddings and assigns a feature label to each token (see Section~\ref{sec:def}).
Given an input $d_i$ and its demonstration $X_i$ (obtained by the DI module), the embedding of a token $w_{ij}$ in $d_i$ is
given by $h_{w_{ij}} = \Mlang(w_{ij}, d_i, X_i)$.
Also, the probability that the token $w_{ij}$ is assigned the label feature $l_{ij}$ is given by
$p_{w_{ij}l_{ij}} = \Mclass(l_{ij} | h_{w_{ij}})$.
Readers are referred to \cite{yang2020simple} for more implementation details on $\Mclass$ and $\Mlang$.

\begin{figure}[!]
\centering
\includegraphics[width=0.37\textwidth]{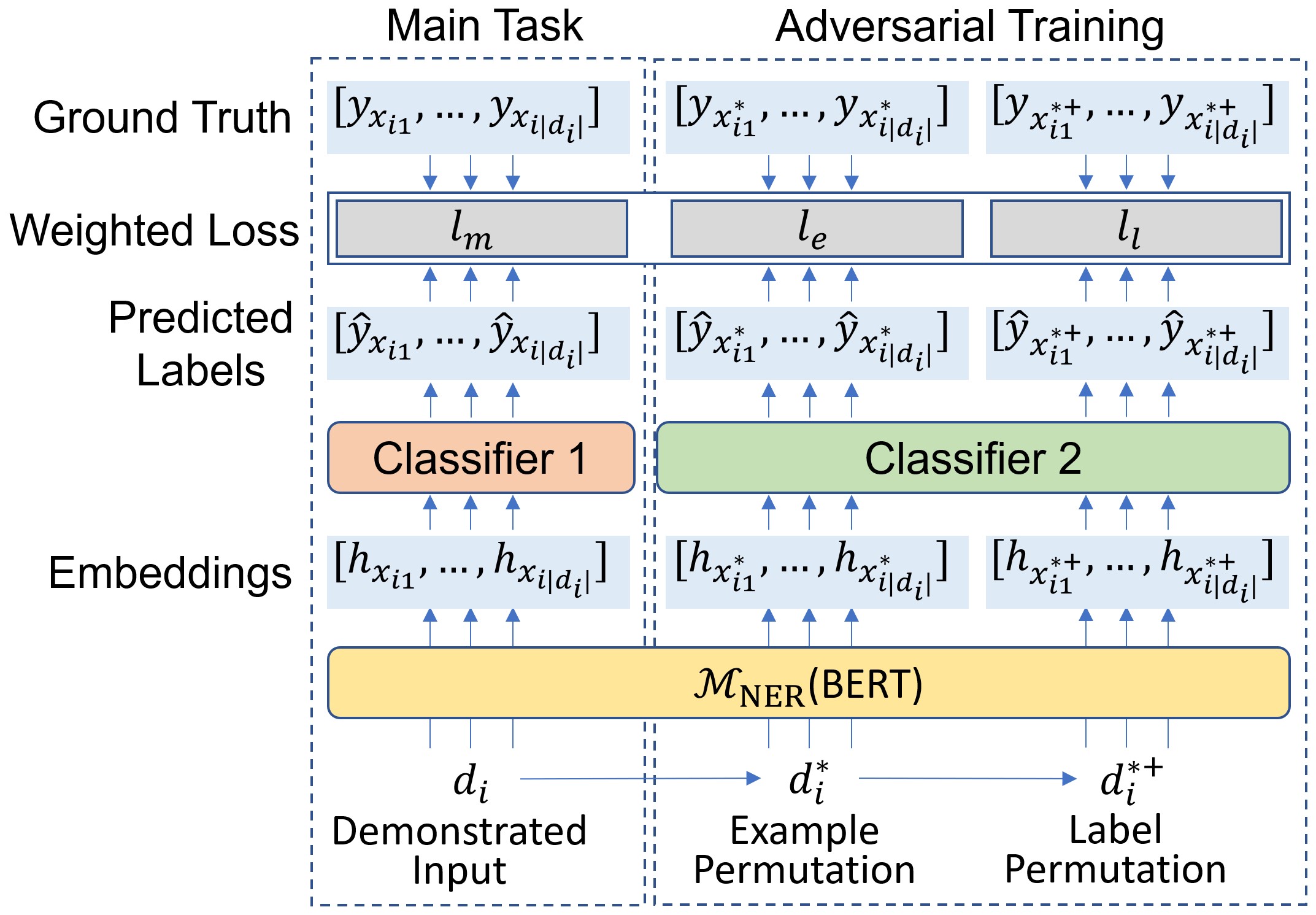}
\caption{Model Training with ADL.}
\label{fig:training_with_game_playing}
\end{figure}

Since the language model $\Mlang$ is fine-tuned from a general-purpose pre-trained model (such as BERT),
The embedding $h_{w_{ij}}$ given by $\Mlang$ could be strongly influenced by the pre-trained model, leading to insufficient 
recognition of the contexts given by the demonstration $X_i$. 
To help the model pay enough attention to the demonstrations, we propose ADL. 
The idea is to modify the annotation rules in creating training samples with perturbed feature labels of tokens. 
With ADL, the model must refer to the demonstration to figure out the annotation rules.
This effectively forces the model to consider the demonstration when making a labeling decision.
Figure~\ref{fig:training_with_game_playing} shows the architecture of model training with ADL.
It consists of two components: (1) A {\it main task} (left component) that follows a traditional NER model training process.
(2) An {\it adversarial training module} (right) that is designed to fine-tune $\Mlang$ so that it is {\it demo-attentive}.
The adversarial training module is used only during training to update the hidden parameters of $\Mlang$;
When labeling a previously unseen input, only the main task module is used.

\begin{figure}[!]
    \centering
    \includegraphics[width=0.5\textwidth]{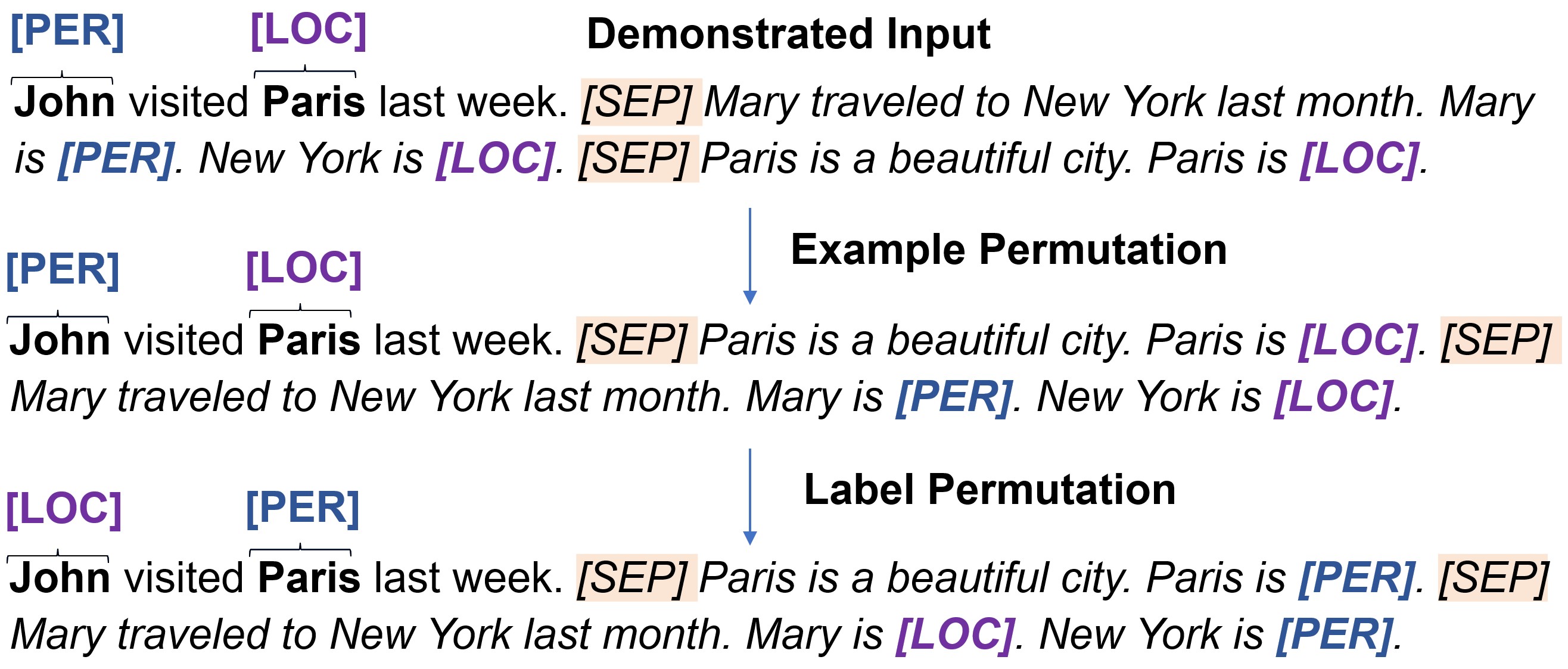}
    \caption{An example of {\it example permutation} and {\it label permutation} in ADL.}
    \label{fig:ExampleAndLabelPermutaion}
\end{figure}

Figure~\ref{fig:ExampleAndLabelPermutaion} shows the process of creating an adversarial demonstration.
Specifically, we first apply example permutation to shuffle the examples within a demonstration. 
This forces the model to refer to different parts of the demonstration to find the useful knowledge.
Secondly, we perform label permutation to change the annotation rule.
For example, in Figure~\ref{fig:ExampleAndLabelPermutaion}, one of the examples in the demonstration reads ``{\it Mary traveled to New York last month. Mary is [PER]. New York is [LOC].}''
With label permutation, we swap the two labels: ``[LOC]'' and ``[PER]'', and the example becomes ``{\it Mary traveled to New York last month. Mary is [LOC]. New York is [PER]}''.
After the permutation, we train the classifier to label according to the new rule, and expect it to label the input ``John'' as ``[LOC]'' and ``Paris'' as ``[PER]''.


From the architecture (Figure~\ref{fig:training_with_game_playing}), we see that there are three parts of loss involved in the training: main task (denoted by $l_m$), example permutation ($l_e$), and label permutation ($l_l$).
We combine these losses by $\mathit{loss} = \alpha l_{m} + (1-\alpha)((1-\beta) l_{e} + \beta l_{l})$, where $\alpha$ is a parameter that controls the relative weight of the main task against that of the adversarial training, and $\beta$ controls the relative weight of regular annotation rules against that of permuted annotation rules within the adversarial training.
The two parameters are adjusted so that the model is applying knowledge obtained from the pre-trained language model
while at the same time paying sufficient attention to the adversarial examples.
The parameter values are determined using grid search.

\section{Experiments}

In this section we experimentally evaluate \method, particularly on the effectiveness of our proposed dual similarity measure and adversarial demonstration training.

\subsection{Datasets and Training}
Our evaluation is based on the following two NER datasets:

\noindent$\bullet$ {\bf CoNLL03 dataset (CoNLL)}~\cite{sang2003introduction} is an NER dataset in the general domain. 
The dataset contains 22K sentences with 4 entity types (features): location, person, organization, and miscellaneous. 

\noindent$\bullet$ {\bf German legal dataset (GL)}~\cite{leitner2020dataset} is a collection of 750 legal documents from 7 courts. The dataset contains 67K sentences with 7 entity types (features): person, location, organization, legal norm, case-by-case regulation, court decision and legal literature.

\noindent$\bullet$ {\bf MIT restaurant review dataset (MIT)}
~\cite{ushio-camacho-collados-2021-ner} contains 9K sentences with 8 entity types (features): rating, amenity, location, restaurant name, price, hours, dish and cuisine. 


The 
datasets exhibit different characteristics.
The sentences in CoNLL are generally shorter and sometimes incomplete, while the sentences in GL are generally 
longer and more complex as they were extracted from court judgments.
The sentences in MIT restaurant reviews are generally shorter than CoNLL, but the average length of markups is longer than CoNLL.
The datasets were divided into training/validation/test sets by the providers.
In the experiments, we adopted few-shot settings.
To construct our few-shot training set, we follow the strategy in \cite{ma_template_free_2022} and sample $k \in \{5, 10, 20\}$ instances 
for each entity type from the original training set.
The validation set is constructed by randomly sampling a total of $k\times|\mathcal{F}|$ instances from the original validation set irrespective of the entity types.
All the instances in the original test sets are used for testing.

\subsection{Other Methods}
We compare \method\ against 12 other methods. We classify all methods into the following 6 categories. 
The methods and their categories are shown in 
Table~\ref{tab:overall_f1_std}.
\newline\noindent$\bullet$ {\bf Basic NER}: 
Basic methods that do not apply data augmentation or prompting. [BERT+CRF~\cite{lee2022good}, NNShot, and StructShot \cite{yang2020simple}].
\newline\noindent$\bullet$ {\bf Implicit prompt tuning}: 
Instead of providing natural language text in the input as the prompt, which is subsequently embedded into vectors, implicit prompt tuning provides additional vectors to the input and hidden layers as the prompt [P-tuning v2 \cite{liu2021p}].
\newline\noindent$\bullet$ {\bf Semantic label}: These methods leverage the semantic information of entity types. In EntLM and EntLM+Struct~\cite{ma_template_free_2022}, the model is asked to output entities that belong to the same type as a given token $t$. For instance, given the token ``Microsoft'', instead of predicting ``[ORG]'', the model may predict ``Google'' as they both belong to the same entity type. For Dual BERT~\cite{ma2022label}, the model aims to maximize the similarity between a given token $t$ and the name of the entity type it belongs to. For example, the similarity between the token ``Microsoft'' and the entity type name ``organization'' is maximized.
\newline\noindent$\bullet$ {\bf Data augmentation}: The training data is augmented by replacement and shuffling [BERT+CRF and StructShot~\cite{dai2020analysis}].
\newline\noindent$\bullet$ {\bf Fixed demonstration}: The best demonstration as determined from the validation set is selected and used for all inputs
[BERT+CRF and StructShot~\cite{lee2022good}].
\newline\noindent$\bullet$ {\bf Dynamic demonstration}: The demonstration is selected based on each input [StructShot~\cite{gao2021making}].

We also compare \method\ with two  
ablated versions, namely,
\method$\neg$ADL (no adversarial demonstration learning), and
\method$\neg$DS (based on semantic similarity only instead of dual similarity).

\subsection{Results}

In Table~\ref{tab:overall_f1_std} we report the F1 scores 
and standard deviations of all methods on CoNLL, GL, and MIT 
datasets under \{5,10,20\}-shot settings.
The results are averaged over 20 experiment runs. 
In the following discussion, we use nicknames to refer to the methods (shown in the table) to save space.
From the table, we make the following observations:



\noindent$\bullet$ For the GL dataset (legal text), two of the basic NER methods, namely, BasicBC and BasicSS,
outperform PTune and all three methods of the semantic label category.
Although BasicBC and BasicSS do not use additional knowledge such as the implicit prompt, they exhibit better performance as they utilize the Viterbi decoder to infer labels while taking label dependencies into account.
The label dependencies are more notable when the annotation spans are longer, which is often the case for GL 
as legal documents typically have longer sentences. 
\newline\noindent$\bullet$ For the CoNLL and the MIT datasets,  
PTune outperforms BasicBC and BasicSS, showing that implicit prompts are useful in improving NER for regular shorter sentences.
\newline\noindent$\bullet$ Data augmentation (DA) is effective for  
all datasets.
This can be seen by comparing the F1 scores of the basic methods against the same methods with DA.
The improvement is especially substantial in the resource-scarce, 5-shot scenario;
For example, at $k$ = 5, DA significantly improves the F1 score from 15.09\% (BasicBC) to 33.23\% (DABC) for the CoNLL dataset.
\newline\noindent$\bullet$ Comparing data augmentation (DABC, DASS) with fixed demonstration (FDBC, FDSS), we observe that fixed demonstration performs better for CoNLL dataset, while data augmentation leads to a better performance for GL dataset.
The reason is that for GL, the annotation task is much more complex --- annotations are done on court judgments which are written by domain experts, and there are more entity types to label.
Having a fixed demonstration for all input samples does not improve the performance by much for challenging NER tasks.
For MIT, both data augmentation and fixed demonstration perform similarly while its entity complexity is between that of CoNLL and GL in term of length of markups and the knowledge for entity type.
\newline\noindent$\bullet$ Compared with fixed demonstration, dynamic demonstration brings further improvement by selecting a better demonstration for each input. This is evidenced by the fact that DDSS performs better
than FDSS for both datasets and all $k$ settings.
\newline\noindent$\bullet$ 
\method\ achieves the best performance among all methods.
Recall that \method\ employs dual similarity and adversarial demonstration training on top of StructShot with dynamic demonstration (DDSS).
Comparing \method\ with DDSS, we see that our \method\ give much better performances across the board.
For example, for GL, \method\ improves the F1 score from 35.73\% to 41.92\% under the 5-shot setting.
This shows that the two techniques we employ, namely, dual similarity and adversarial demonstration learning 
are highly effective in selecting the best demonstration and helping the model effectively utilize the knowledge embedded in the demonstration.
\newline\noindent$\bullet$ 
We see a performance drop when \method\ is regressed to either of the ablated versions (\method$\neg$ADL or  \method$\neg$DS).
Yet, the performance of either ablated version is better than the basic dynamic demonstration method (DDSS). 
This shows that both dual similarity and adversarial demonstration learning contribute to the improvement of the tagger, and that the two proposed techniques complement each other well.

\begin{table*}
	\centering
	\caption{Average F1 scores (in \%) and standard deviations of 20 experiment runs on CoNLL, GL and MIT datasets. 
	}
	\resizebox{.95\linewidth}{!}{
		\begin{tabular}{llrrrrrrrrrr}
			\toprule
			\multirow{2}{*}{Category}  & \multirow{2}{*}{Method} & \multicolumn{3}{c}{CoNLL} & \multicolumn{3}{c}{GL} & \multicolumn{3}{c}{MIT} \\ \cmidrule(lr){3-5} \cmidrule(lr){6-8} \cmidrule(lr){9-11}
			& & \multicolumn{1}{c}{$k$ = 5}  & \multicolumn{1}{c}{10}    & \multicolumn{1}{c}{20}   & \multicolumn{1}{c}{5}     & \multicolumn{1}{c}{10}    & \multicolumn{1}{c}{20}  & \multicolumn{1}{c}{5}     & \multicolumn{1}{c}{10}    & \multicolumn{1}{c}{20}    \\ \midrule
			\multirow{3}{*}{Basic NER} & BERT+CRF (BasicBC)                                       & 15.09 {\small $\pm$ 9.65} & 42.35 {\small $\pm$ 8.04}  & 59.17 {\small $\pm$ 5.32}  & 33.15 {\small $\pm$ 4.92}  & 45.29 {\small $\pm$ 4.12}  & 55.17 {\small $\pm$ 2.20}   &27.67{\small $\pm$3.35} &	42.85{\small $\pm$3.06} &	53.48{\small $\pm$2.24} \\ \cmidrule(lr){2-11}
			& StructShot (BasicSS)                                    & 23.00 {\small $\pm$ 10.48}  & 44.20 {\small $\pm$ 7.79}  & 58.01 {\small $\pm$ 2.95}  & 33.65 {\small $\pm$ 5.40}  & 46.39 {\small $\pm$ 4.74}  & 54.87 {\small $\pm$ 2.76}   &24.62{\small $\pm$3.71} &	39.78{\small $\pm$3.01} &	52.67{\small $\pm$1.98} \\ \cmidrule(lr){2-11}
			& NNShot (BasicNNS)                                       & 19.93 {\small $\pm$ 4.06}  & 41.89 {\small $\pm$ 7.61}  & 57.55 {\small $\pm$ 4.11}  & 27.66 {\small $\pm$ 6.12}  & 42.24 {\small $\pm$ 2.95}  & 51.73 {\small $\pm$ 2.86}   &26.63{\small $\pm$3.64} &	40.87{\small $\pm$3.27} &	52.96{\small $\pm$2.27} \\ \midrule
			\multicolumn{1}{l}{Implicit prompt tuning} & P-tuning v2 (PTune)                                  & 25.92 {\small $\pm$ 6.57}  & 51.16 {\small $\pm$ 6.97}  & 63.26 {\small $\pm$ 4.92}  & 28.66 {\small $\pm$ 5.29}  & 42.32 {\small $\pm$ 3.23}  & 53.73 {\small $\pm$ 2.69}   &30.54{\small $\pm$3.91} &	43.40{\small $\pm$2.84} &	53.55{\small $\pm$2.08} \\ \midrule
			\multirow{3}{*}{Semantic label} & EntLM                                          & 22.91 {\small $\pm$ 7.15}  & 41.18 {\small $\pm$ 8.89}  & 57.46 {\small $\pm$ 3.51}  & 22.17 {\small $\pm$ 6.12}  & 34.76 {\small $\pm$ 3.61}  & 46.11 {\small $\pm$ 2.98}   &28.43{\small $\pm$4.46} &	40.67{\small $\pm$2.74} &	51.86{\small $\pm$2.02} \\ \cmidrule(lr){2-11}
			& EntLM+Struct (EntLM+S)                                  & 25.49 {\small $\pm$ 5.17}  & 42.07 {\small $\pm$ 9.52}  & 60.43 {\small $\pm$ 4.95}  & 31.01 {\small $\pm$ 4.36}  & 39.24 {\small $\pm$ 2.02}  & 45.29 {\small $\pm$ 2.48}   &29.04{\small $\pm$4.13} &	41.36{\small $\pm$2.20} &	52.15{\small $\pm$2.50} \\ \cmidrule(lr){2-11}
			& Dual BERT (D-BERT)                                    & 15.81 {\small $\pm$ 10.35} & 43.30 {\small $\pm$ 6.66}  & 57.89 {\small $\pm$ 4.73} & 31.40 {\small $\pm$ 5.32}  & 44.39 {\small $\pm$ 3.01} & 53.95 {\small $\pm$ 2.16}  &26.15{\small $\pm$4.30} &	39.48{\small $\pm$3.49} &	52.16{\small $\pm$1.88} \\ \midrule
			\multirow{2}[0]{*}{Data augmentation (DA)} & BERT+CRF (DABC)                    & 33.23 {\small $\pm$ 11.02} & 55.01 {\small $\pm$ 6.00} & 64.52 {\small $\pm$ 3.82} & 35.36 {\small $\pm$ 6.48} & 47.59 {\small $\pm$ 3.46} & 56.50 {\small $\pm$ 2.45}  &32.33{\small $\pm$3.88} &	47.41{\small $\pm$2.61} &	56.07{\small $\pm$1.56} \\ \cmidrule(lr){2-11}
			& StructShot (DASS)                 & 32.58 {\small $\pm$ 8.47} & 52.68 {\small $\pm$ 5.59} & 63.80 {\small $\pm$ 3.94} & 39.32 {\small $\pm$ 4.99} & 50.81 {\small $\pm$ 3.33} & 58.12 {\small $\pm$ 2.88}  &32.73{\small $\pm$3.77} &	47.55{\small $\pm$3.70} &	57.34{\small $\pm$1.62} \\ \midrule
			\multirow{2}{*}{Fixed demonstration (FD)} & BERT+CRF (FDBC)            & 45.14 {\small $\pm$ 9.85} & 59.60 {\small $\pm$ 5.19}  & 66.34 {\small $\pm$ 3.77} & 29.81 {\small $\pm$ 4.74} & 43.22 {\small $\pm$ 3.35} & 54.88 {\small $\pm$ 2.53}  &31.70{\small $\pm$5.14} &	44.60{\small $\pm$2.68} &	53.99{\small $\pm$1.50} \\ \cmidrule(lr){2-11}
			& StructShot (FDSS)                                & 43.83 {\small $\pm$ 9.00} & 61.34 {\small $\pm$ 4.95} & 68.03 {\small $\pm$ 4.23} & 34.24 {\small $\pm$ 5.12} & 48.33 {\small $\pm$ 2.49} & 56.69 {\small $\pm$ 2.49}  &34.49{\small $\pm$5.49} &	47.46{\small $\pm$3.31} &	56.74{\small $\pm$1.89} \\ \midrule
			\multirow{4}{*}{\begin{tabular}{@{}l@{}}Dynamic \\ demonstration (DD) \end{tabular} } & StructShot (DDSS)       & 47.84 {\small $\pm$ 6.76} & 62.69 {\small $\pm$ 4.76} & 70.39 {\small $\pm$ 2.77}  & 35.73 {\small $\pm$ 5.37} & 49.37 {\small $\pm$ 4.09} & 58.08 {\small $\pm$ 3.24}   &35.75{\small $\pm$4.07} &	49.59{\small $\pm$3.24} &	58.85{\small $\pm$1.90} \\ \cmidrule(lr){2-11}
			& \textbf{\method}    & \textbf{51.78} {\small $\pm$ 6.28} & \textbf{64.88} {\small $\pm$ 3.62} & \textbf{72.13} {\small $\pm$ 2.30}  & \textbf{41.92} {\small $\pm$ 3.66} & \textbf{53.63} {\small $\pm$ 2.01} & \textbf{60.99} {\small $\pm$ 2.85}   &\textbf{40.10}{\small $\pm$4.60} &	\textbf{51.65}{\small $\pm$2.89} &	\underline{59.57}{\small $\pm$2.03} \\ \cmidrule(lr){2-11}\morecmidrules\cmidrule(lr){2-11}
			& [Ablated] \method$\neg$ADL                 & 48.96 {\small $\pm$ 6.98} & \underline{64.70} {\small $\pm$ 4.45}  & \underline{71.64} {\small $\pm$ 3.02}  & 37.47 {\small $\pm$ 4.90} & 51.07 {\small $\pm$ 3.36} & 58.86 {\small $\pm$ 3.51}   &37.19{\small $\pm$4.96} &	\underline{51.35}{\small $\pm$3.26} &	\textbf{59.93}{\small $\pm$2.03} \\ \cmidrule(lr){2-11}
			& [Ablated] \method$\neg$DS                       & \underline{49.45} {\small $\pm$ 5.43} & 63.81 {\small $\pm$ 4.30} & 69.80 {\small $\pm$ 3.51}  & \underline{40.32} {\small $\pm$ 3.13} & \underline{52.74} {\small $\pm$ 2.43} & \underline{60.33} {\small $\pm$ 2.58}  &\underline{37.95}{\small $\pm$4.77} &	49.71{\small $\pm$3.49} &	58.68{\small $\pm$1.97} \\ \bottomrule 
		\end{tabular}
	} 
	\label{tab:overall_f1_std}
\end{table*}

\subsection{Feature Similarity Predictor}
In Section~\ref{sec:adell}, we show how our dual similarity function $S$ integrates both feature similarity $\sfe$
and semantic similarity $\sse$ (see Eq.~\ref{eq:dual-sim} and Fig.~\ref{fig:SimilarityPredictior}).
In particular, we train a predictor that outputs feature similarity score $\sfe(d_i, d_j)$ that estimates the feature 
Jaccard similarity, $\mathit{FJ}(d_i, d_j)$ of two given input text pieces $d_i$ and $d_j$ (Eq.~\ref{eq:fj}).
The estimates are then used to help rank demonstrative examples in constructing demonstrations. 
In this section we evaluate the effectiveness of our predictor (i.e., $\sfe$).
We consider three accuracy metrics:
\newline\noindent$\bullet$ {\bf Binary accuracy}: 
For each input $d$ in the test set, we pick two examples $d_i$ and $d_j$ from the original training set such that $d$ and $d_i$ share some common features
but $d$ and $d_j$ do not.
That is, 
$\mathit{FJ}(d, d_i) > 0$ and $\mathit{FJ}(d, d_j)=0$. 
We compute the probability of $\sfe(d, d_i) > \sfe(d, d_j)$
by repeating the procedure 10,000 times with different
pairs of $(d_i, d_j)$. 
\newline\noindent$\bullet$ {\bf Ranking accuracy}: 
The pair $(d_i, d_j)$ are picked such that $\mathit{FJ}(d, d_i) > \mathit{FJ}(d, d_j) > 0$.
The ranking accuracy of $\sfe$ is measured by the probability of $\sfe(d, d_i) > \sfe(d, d_j)$.
\newline\noindent$\bullet$ {\bf Pearson correlation}: For each input $d$ in the test set, we pick an example $d_i$ and compute the correlation between $\mathit{FJ}(d, d_i)$ and $\sfe(d, d_i)$ for all $d$'s and $d_i$'s.

Table~\ref{tab:retriever_accuracy} shows the accuracy results of the predictor ($\sfe$) that was trained under \{5, 10, 20\}-shot settings.
From the table, we see that $\sfe$ produces good accuracies for both CoNLL and GL datasets, demonstrating that the feature similarity predictor can well predict the Jaccard similarity between input text pieces.
This is also corroborated by the medium-to-high correlation between $\sfe$ and $\mathit{FJ}$, especially for the more complex and feature-rich legal text (GL).
The ranking accuracy results also show that the feature similarity predictor adequately ranks demonstrative examples with regard to their feature similarity.

Many existing demonstrative learning methods use only semantic similarity (e.g., $\sse$) to rank demonstrative examples.
In Table~\ref{tab:retriever_accuracy}, we show the accuracy results if $\sse$ were used to estimate the feature Jaccard similarity
as an interesting reference.
From the table, we see that $\sse$ has no correlation with feature similarity for CoNLL dataset and negative correlation for the GL dataset. 
This shows that the traditional way of using semantic similarity alone in ranking demonstration examples largely ignores 
(or is in conflict with) feature similarity. 
Moreover, the ranking accuracies for $\sse$ are very poor, and the Pearson correlation between $\sse$ and feature similarity is negative
, showing that considering examples' semantic alone 
fails to rank demonstration examples correctly w.r.t. their features. 
These shortcomings are remedied by our dual similarity measure, which considers both $\sse$ and $\sfe$.

\begin{table}[H]
  \centering
  \caption{Similarity predictor accuracies.}
  \resizebox{\linewidth}{!}{
  \setlength{\extrarowheight}{3pt}
  \begin{tabular}{lrrrr|rrrrr}
    \hline
    \multirow{3}{*}{Metrics} &\multicolumn{4}{c|}{CoNLL} & \multicolumn{4}{c}{GL}         \\ \cline{2-5} \cline{6-9}
                             & \multicolumn{3}{c}{$\sfe$} &\multirow{2}{*}{$\sse$} & \multicolumn{3}{c}{$\sfe$} &\multirow{2}{*}{$\sse$}  \\ \cline{2-4} \cline{6-8}
                              & \multicolumn{1}{c}{$k$ = 5}     & \multicolumn{1}{c}{10}     & \multicolumn{1}{c}{20}    &    & \multicolumn{1}{c}{5}     & \multicolumn{1}{c}{10}     & \multicolumn{1}{c}{20}   & \\ \hline
    Binary accuracy (in \%) & 64.02 & 70.65 & 74.01 & 52.93 & 88.08 & 90.04 & 92.47 & 34.18 \\ \cline{2-5} \cline{6-9}
    Ranking accuracy (in \%) & 69.40 & 70.50 & 72.85 & 39.44 & 84.67 & 86.65 & 88.37 & 18.02 \\ \cline{2-5} \cline{6-9}
    Pearson correlation      & 0.3912 & 0.4823 & 0.5533 & -0.0519 & 0.7582 & 0.8070 & 0.8462 & -0.5351 \\ \hline
  \end{tabular}
  } 
\label{tab:retriever_accuracy}
\end{table}


\subsection{Effectiveness of Adversarial Demonstration Learning}
\begin{figure}[h!]
	\centering
	\subfigure[Without ADL]{
		\label{fig:lp_basic_demo} 
		\includegraphics[width=0.2\textwidth]{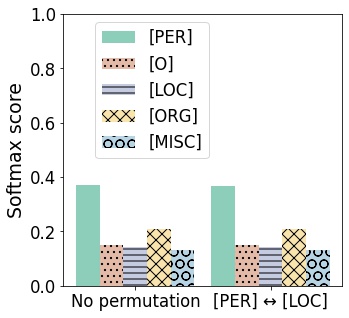}}
	\subfigure[With ADL]{
		\label{fig:lp_ad_demo} 
		\includegraphics[width=0.2\textwidth]{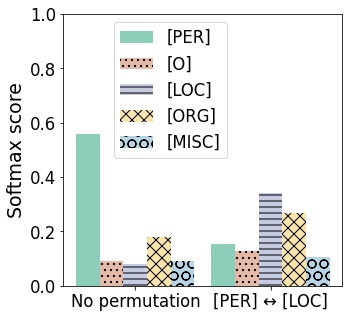}}
	\caption{Average label scores of [PER] tokens in the test set.}
	\label{fig:CosineSimilarityEmbedding} 
\end{figure}

We study how successful adversarial demonstration learning (ADL) helps guide an NER model to pay sufficient 
attention to the demonstrations, thus making demonstration learning more effective. 
First, we train a regular NER model without applying ADL. 
We then apply the model to the inputs in the test set. For each input, demonstrations are incorporated as 
discussed in Section~\ref{sec:adell} and Figure~\ref{fig:framework_dynamicdemonstration_learning}(c). 
Now, for each [PER] token in the inputs, we note the label scores given by the NER model for each label type.
The left bar chart in Figure~\ref{fig:lp_basic_demo} (labeled ``no permutation'') shows
the averages of such scores (for different label types). 
For example, the chart shows that of all the [PER] tokens, the average label scores [PER] and [LOC] are 0.37 and 0.15, respectively.
Next, we intentionally fool the NER model by providing mis-labeled demonstrations. 
Specifically, we change all [PER] labels in all demonstrative examples to [LOC] and vice versa. 
We repeat the testing and the resulting NER label scores are plotted in the right chart of
Figure~\ref{fig:lp_basic_demo}.
Comparing the left and right charts of Figure~\ref{fig:lp_basic_demo}, we see that label scores 
of [PER] and [LOC]) did not change by much. 
This shows that the NER model was not influenced much by the demonstrations used in testing, but relied 
mostly on the knowledge learned during training to make its predictions.

We repeat the experiment but this time the NER model is trained with ADL. 
The resulting label scores are shown in Fig.~\ref{fig:lp_ad_demo}.
We see that the NER model trained with ADL is much more accurate, as evidenced by the much
higher score of the [PER] label (for [PER] tokens) shown in the left chart of the figure.
Moreover, the labels' scores are drastically changed if we intentionally swap the [PER] and [LOC] labels
in the demonstrative examples used in testing. This shows that the model trained with ADL pays attentions
to the demonstration given (and thus were influenced by the mis-labeled demonstrations).
Even though we only show the results of swapping [PER] and [LOC] labels in testing, 
similar conclusions are drawn when we repeat the experiments and swap other label pairs.
In conclusion, our method of adversarial demonstration learning is effective for
enhancing the NER model by amplifying its attention to demonstrations.

\begin{figure}
  \centering

  \subfigure[Without demonstration learning (BasicSS)]{
    \label{fig:zhangwei_lijie_no_demo} 
    \includegraphics[width=0.47\textwidth]{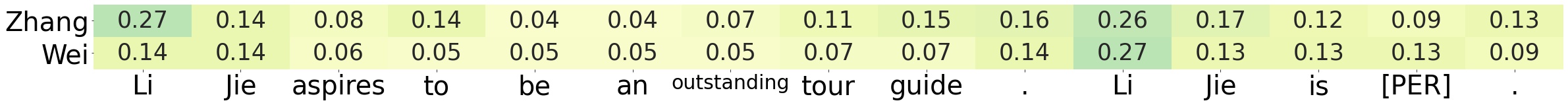}}

  \subfigure[With dynamic demonstration learning (DDSS)]{
    \label{fig:zhangwei_lijie_no_adversarial_demo} 
    \includegraphics[width=0.47\textwidth]{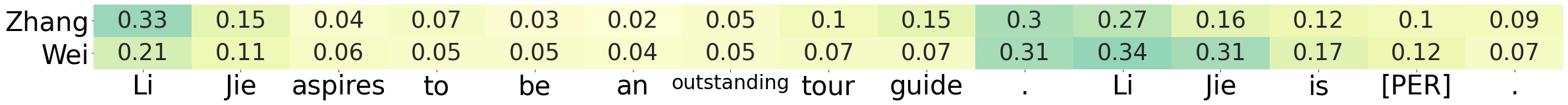}}

  \subfigure[With ADL (\method)]{
    \label{fig:zhangwei_lijie_with_adversarial_demo} 
    \includegraphics[width=0.47\textwidth]{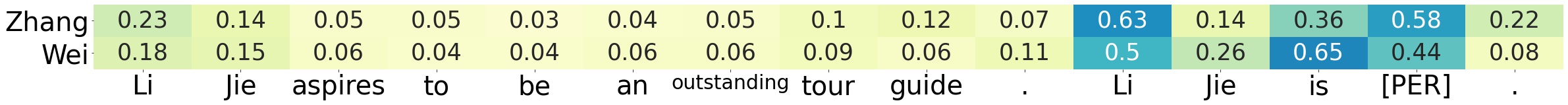}}
  \caption{Maximum attention score between ``\textbf{Zhang Wei}'' and each word in the example sentence. The input is: ``\textbf{Zhang Wei} went to visit the Eiffel Tower.'' The demonstrative example is: ``Li Jie aspires to be an outstanding tour guide. Li Jie is \text{[PER]}.''}
  \label{fig:Attention1} 

\end{figure}

We further study the influence of ADL on the attention mechanisms of the language model.
We train three NER models using three different approaches: (1) without demonstration learning (BasicSS), (2) with dynamic demonstration learning (DDSS), and (3) with adversarial demonstration learning (\method). 
We then compute the maximum attention scores (across all layers and heads in BERT) between the entity in the input and each word in the demonstrative example.
For example,
the attention scores are visualized as heatmaps in Figure~\ref{fig:Attention1} for the input: ``{\it Zhang Wei went to visit the Eiffel Tower.}'' and the demonstrative example: ``{\it Li Jie aspires to be an outstanding tour guide. Li Jie is \text{[PER]}.}''

From the figure, we see that the model trained with demonstration learning (DDSS) gives better attention than that
trained with no demonstration (BasicSS). This shows how demonstration learning helps improve model performance.
More interestingly, the model trained with ADL (\method) gives high attention scores to the tokens ``Li Jie'' in the demonstration for the input words ``Zhang Wei''. 
These results show that our method is effectiveness in directing models to focus towards the demonstrations.


We have conducted extensive experiments to further evaluate our proposed method \method. Due to space constraints, some of the experimental results are omitted in the paper. 
For example,
we have conducted case studies to examine the performance of our example retrieved methods and the influence of ADL on the attention mechanism respectively.
Furthermore, we  further compare \method\ against the baseline method DDSS across various demonstration construction scenarios. The results show that our dual similarity function and ADL improve  demonstration learning in all tested scenarios.
Readers are referred to the supplementary materials for the additional experiment results and discussions.


\section{Conclusion}

In this paper we investigate techniques that improve 
demonstration learning. 
We identify two important issues in traditional demonstration learning methods: First, the choice of demonstrative
examples relies mostly on semantic similarity between an input example and the demonstrative examples. 
The example used as demonstration fails to match the input in terms of the features they contain. 
Secondly, we observe that in many cases, the NER model trained does not pay enough attention to the demonstrative
examples, which lowers the effectiveness of demonstration learning. 
To tackle these issues, we propose \method, which employs a dual similarity measure and an adversarial demonstration learning (ADL) strategy. 
Specifically, our dual similarity function uses a feature similarity predictor to evaluate how similar two examples are. 
Through extensive experiments, we show that the dual similarity measure is able to capture examples' feature-based similarity effectively. This improves the quality of the demonstration constructed in the learning process.
Also, we show that ADL can effectively direct the model's attention to the demonstrative examples. 
Our experiment results show that \method\ outperforms existing methods and is thus highly effective in
enhancing demonstration learning.
\newpage
\bibliographystyle{named}
\bibliography{demoNER}

\end{document}